\documentclass[letterpaper, 10 pt, conference]{ieeeconf}  % Comment this line out if you need a4paper

\IEEEoverridecommandlockouts                              % This command is only needed if 
\usepackage{graphicx} %include graphics

\usepackage{tabularx} %for tables
\usepackage{booktabs}
\usepackage{hyperref}
\usepackage{caption}
\usepackage{multicol}
\usepackage{tikz}
\usepackage{tikz-3dplot}
\usepackage{arydshln}
\usetikzlibrary{positioning,calc,patterns,angles,quotes}
\usetikzlibrary{quotes,angles}
\usepackage{subfig}
\usepackage[space]{cite}

\newcommand{\fref}[1]{Figure~\ref{#1}}
\newcommand{\tref}[1]{Table~\ref{#1}}
\newcommand{\sref}[1]{Section~\ref{#1}}
\newcommand{\eref}[1]{Equation~(\ref{#1})}

\usepackage{xcolor}

\title{\LARGE \bf A System-driven Automatic Ground Truth Generation Method for DL Inner-City Driving Corridor Detectors}
\author{Jona Ruthardt$^{1}$, Thomas Michalke$^{1*}$% <-this % stops a space
\thanks{$^{1}$Robert Bosch GmbH,
Robert-Bosch-Platz 1, 
70839 Gerlingen-Schillerh\"ohe, Germany
        {\tt\small  Firstname.Lastname@de.bosch.com}}%
\thanks{*contact author}
}
\begin{document}
\maketitle
\thispagestyle{empty}
\pagestyle{empty}
%
%%%%%%%%%%%%%%%%%%%%%%%%%%%%%%%%%%%%%%%%%%%%%%%%%%%%%%%%%%%%%%%%%%%%%%%%%%%%%%%%
\begin{abstract}
Data-driven perception approaches are well-established in automated driving systems. In many fields even super-human performance is reached. Unlike prediction and planning approaches, mainly supervised learning algorithms are used for the perception domain. Therefore, a major remaining challenge is the efficient generation of ground truth data. As perception modules are positioned close to the sensor, they typically run on raw sensor data of high bandwidth. Due to that, the generation of ground truth labels typically causes a significant manual effort, which leads to high costs for the labelling itself and the necessary quality control. In this contribution, we propose an automatic labeling approach for semantic segmentation of the drivable ego corridor that reduces the manual effort by a factor of 150 and more. The proposed holistic approach could be used in an automated data loop, allowing a continuous improvement of the depending perception modules.
% Schönes Acronym finden, ggf. Thema Data Loop verwursteln
%
\end{abstract}
\begin{keywords}
Machine Learning, Deep Learning, Automated Ground Truth, Data Loop
\end{keywords}
%
%%%%%%%%%%%%%%%%%%%%%%%%%%%%%%%%%%%%%%%%%%%%%%%%%%%%%%%%%%%%%%%%%%%%%%%%%%%%%%%%
\section{Introduction}  Lane detection is an essential part of the perception sub-architecture of any automated driving (AD) or advanced driver assistance system (ADAS). In highway scenarios, reasoning and vehicle control strongly rely on information provided by high-precision (HD) maps and lane markings. Here, HD maps are used as the primary input for lateral control as well as for the selection of relevant objects via map-based lane association.
More recently and with a focus on inner-city scenarios, data-driven approaches have been proposed that detect the freespace (drivable area in a physical sense). The focus of AD approaches in inner city scenarios is less on lane-based driving due to the fact that the lane concept does not fully apply in unmarked, residential urban environments (e.g., a narrow inner-city scenario with rows of parking and oncoming vehicles). In unmarked inner-city environments, human-steered vehicles tend to drive according to the available freespace and not according to a fuzzy lane definition. Hence, representing the unstructured inner-city world by pre-defined HD maps and realizing reasoning on that does not work, as other traffic participants do not behave according to that static representation.

Therefore, in a recent contribution \cite{Michalke2021}, we proposed to rely on a data-driven approach for generating the drivable space that extends the state-of-the-art by adding more semantics to the typically used freespace. More specifically, we proposed the concept of an ''AI ego corridor'' that classifies the corridor the ego vehicle is allowed to use. Instead of a static map, this is the basis for lateral control and for determining behavior-relevant objects, especially in challenging inner-city scenarios.

\begin{figure*}[htb]
	\centering
	\includegraphics[width=0.8\linewidth]{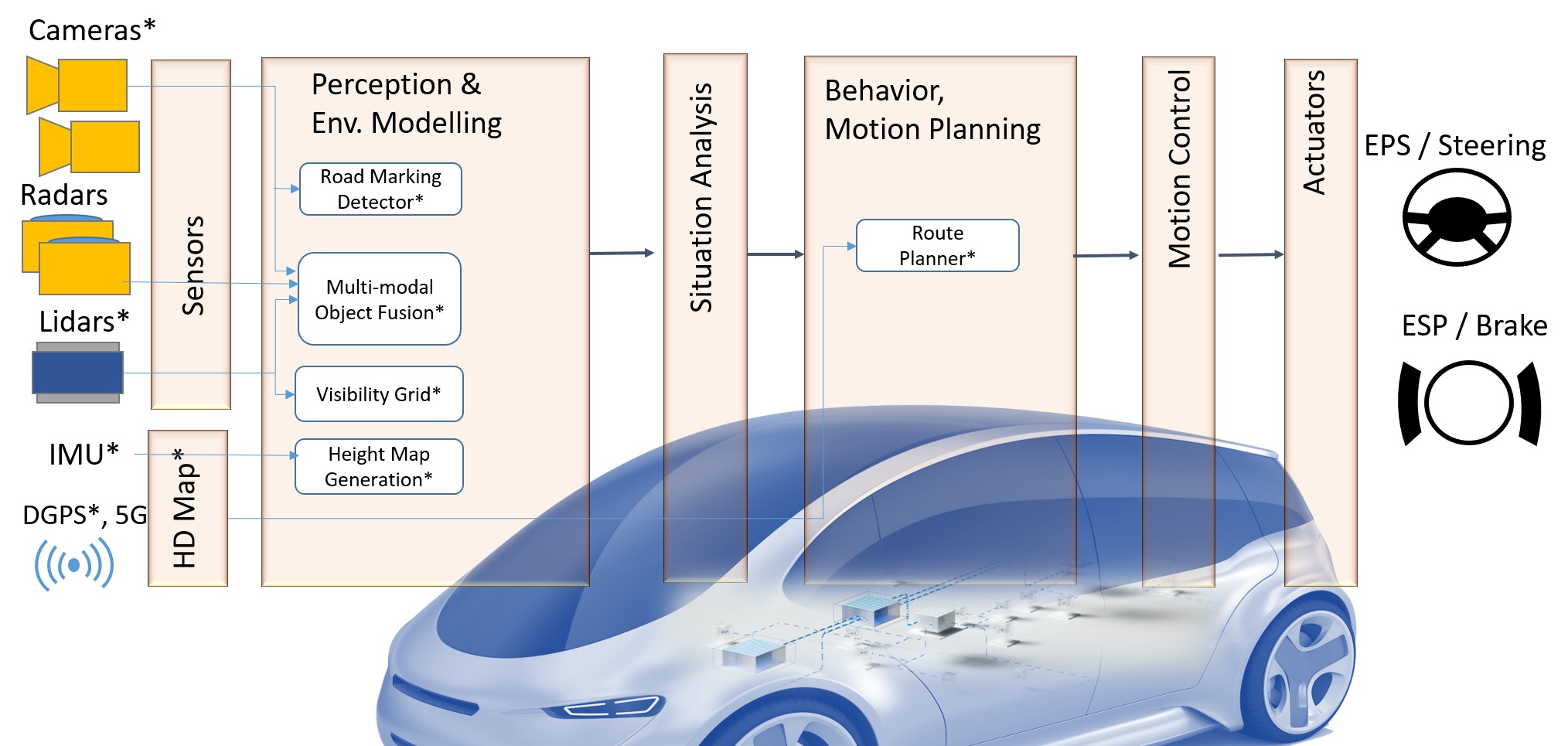}
	\caption{Used AD stack (all modules marked with * contribute data to the automatic GT generation pipeline)}
	\label{fig:ADpipeline}
\end{figure*}

As for all image-based semantic segmentation approaches, we also face the challenge of creating sufficient labelled ground truth (GT) data in terms of quantity as well as quality. In this contribution, we present a pipeline for automated GT generation that is closely coupled to a L3 AD stack running on an AD vehicle featuring a high-performance multi-modal sensor set (see \fref{fig:ADpipeline}). 

The proposed system-driven solution allows for an automatic ground truth generation even in unmarked, complex inner-city scenarios. The reduction in labeling effort is very significant. Given our experiments, a reduction factor in manual labeling time of more than 150 was achieved. 

The remaining paper is structured as follows:
\sref{chap:related_work} discusses related contributions and derives the still unresolved research questions. \sref{chap:method} gives a detailed description of our system-driven approach for automated ground truth generation. In \sref{chap:Experiments}, we apply the approach on a semantic segmentation algorithm for detecting the drivable ego-corridor and evaluate its impact on the manual labeling effort as well as the labeling noise.

\section{Related Work}     \label{chap:related_work}

%\subsection{Lane Detection}

Detecting lanes and thus determining the trajectory to follow is an essential part of AD and has been researched for decades. Consequently, a lot of different approaches have been proposed to tackle this problem \cite{Kumar2015}. The technologies and concepts thereto deployed vary substantially, however \cite{Narote2018}. 

Whereas feature-based (e.g., \cite{Niu2015}) or model-based (e.g., \cite{Lee2002}) methods rely on characteristics within the available input data that can be extracted and evaluated in a rule-based or heuristic framework, data-driven machine learning (ML) approaches require the training of a model using supervisory signals. The drawback of requiring large volumes of typically annotated training and evaluation data for such ML models is often offset by their substantial performance advantage compared to classical approaches \cite{Xing2018}. As a result, the development of data-driven models currently receives much attention from the research community and is a trend that is likely to continue in the next years \cite{Chetan2020}.

Owing to the emerging interest in ML-based lane detection approaches and their demand for large volumes of GT data, a variety of public datasets were created (e.g., \cite{Geiger2013,Scharwaechter2013,Cordts2016,Yu2018}). However, their distribution and complexity of included traffic scenarios, as well as their types of annotations and the number of contained frames, vary substantially \cite{Zhang2021}. Moreover, the utilized (deep) learning techniques also differ in the type of lane representation they output and subsequently the GT labels they require \cite{Tang2021}. Taken together, this often results in a shortage of GT data with suitable characteristics, high-quality annotations, and coverage of a broad spectrum of driving situations, including complex and rare cases, thus constituting one central limitation of current ML approaches. Especially in the case of less common lane representations, as is the case in our approach, obtaining sufficiently large datasets to train robust and accurate models often involves the cumbersome process of manually creating the desired GT annotations.

%\subsection{Automatic Ground Truth Generation}

The problem of having limited or too little GT data for a specific lane detection use case can be approached in different ways. 

In case the GT data is sparse in a particular environment but readily available in another, \emph{transfer learning} can be utilized by fine-tuning a model that was first trained on the more extensive source domain with data from the ultimate target domain. In doing so, the model can generalize from the insights learned in the first step and is less likely to overfit the task-specific training data. Kim et al. \cite{Kim2017}, for example, have employed transfer learning by fine-tuning a general semantic image segmentation model to only extract and segment the left and right lanes in given images.

Another option is to combine the transfer learning methodology with computer-simulated traffic scenarios. By these means, it is possible to train models with data generated in almost arbitrary numbers in simulations where the environment can be exactly observed and altered as desired, and to optionally only fine-tune on task-specific real-world GT data. In lane detection applications, this approach has already proved to be effective and very time efficient (e.g., \cite{Garnett2020, Hu2022}). Even with increasingly photo-realistic simulations techniques (e.g., \cite{Dosovitskiy2017}) and approaches that try to close the simulation-reality-gap (e.g., \cite{Richter2021}), there are still persisting distributional differences between the data generated in simulation and real-world data, however, making it more difficult for models to adapt to the eventual application domain \cite{Ranaweera2021}. 

Over the last years, there has been an increase in publications proposing novel approaches for (automatically) creating large quantities of real-world lane detection and segmentation GT data while requiring no or only little human intervention. Subsequently, larger volumes of training and evaluation data are available that also share the exact properties of the eventual application domain. 

Methods such as \cite{Borkar2010} or \cite{Borkar2012} reduce the ratio of lane markings that have to be labeled to accurately approximate the lane course. This is achieved by creating time-sliced images through composing horizontal image slices of a video sequence which are then manually annotated before being deconstructed into the original input frames again while inferring the lane course from the interpolated labeled marker positions. While these approaches are more efficient and can achieve a good GT quality when compared to fully manually annotated GT data \cite{Borkar2012}, they still require manual labeling efforts.

More sophisticated approaches can forgo manual intervention in the GT creation process altogether by utilizing a-priori knowledge (e.g., maps) or unsupervised methods that solely rely on characteristics of the input data to extract lane properties and augment the GT representations. 

% Types of maps used & unmarked roads 
Properties of roads and lanes, such as the course and position, are stored in most maps, making them a prime source of a-priori knowledge. This is reflected in the literature, where many automatic GT generation frameworks primarily rely on information extracted from maps. The level of detail and type of maps utilized varies broadly, however. HD maps of static objects as they are used in \cite{Behrendt2017} and \cite{Behrendt2019} can be created during good environmental conditions and constitute a detailed and accurate representation of specific environmental entities. This high definition comes at the expense of their creation and maintenance being time- and resource-intensive and them not being broadly available. Therefore, others choose to use less detailed but openly available maps for their systems. Kasmi et al. \cite{Kasmi2020}, for instance, extract the road segments from OpenStreetMap \cite{OpenStreetMap2017} and detect lanes more accurately by matching the hereby presumed course of the road with LiDAR point clouds. While the objective of this work was not directly to create GT data but to detect lanes, the approach could be used with little modifications for automatic GT generation. Only the road skeletons obtained from geographic information system (GIS) maps were used in \cite{Alvarez2014} and augmented with additional assumptions before ultimately being projected as a road surface segmentation mask into the image plane. By adjusting projection parameters in accordance with additional road clues (e.g., vanishing points, surface color and structure, horizon line, etc.) in an unsupervised manner to improve the fit of the map-based segmentation mask and the corresponding camera image, this approach can generate GT data on unmarked roads as well. Due to their reliance on maps that only include the position of lane markers (\cite{Behrendt2017, Behrendt2019}) or on the reflection intensity of LiDAR-based approaches (\cite{Kasmi2020}), other methods can only be deployed on roads with present lane markers. This significantly limits their applicability, especially in urban scenarios where unmarked roads are common.

% Quality of GTs (especially in urban scenarios)
Another aspect that makes GT generation in urban scenarios more challenging is the increased complexity of the environment and encounterable scenarios. Steep de- or inclines, obstructions through dynamic or static objects and dynamic vehicle movements all impact the desired GT annotation. While some methods implicitly \cite{Alvarez2014} or explicitly \cite{Behrendt2019} consider dynamic vehicle position and orientation to ensure accurate annotations, there is no holistic system-driven approach that considers all of the aforementioned scenarios in the GT generation process by utilizing additional sensors and sources of information about the environment. 

% ego-corridor GT representation
Most publications focus on the detection and segmentation of individual lane markers (\cite{Behrendt2017, Behrendt2019}) or on inferring the course of the road from them (\cite{Kasmi2020, Borkar2012}) rather than constructing an ego-lane corridor that is bound by lane markings, road curbs, and road users or obstacles in front. Although it is possible in many cases to create specific GT types by modifying the given representation (e.g., lane markers can be converted to lane spline from which the preceding lane corridor can be constructed), the absence of explicit object and obstacle detection algorithms in these frameworks prevents the GT data to be used for training models utilizable for longitudinal control. With other GT representations such as in \cite{Alvarez2014} or \cite{Behrendt2017}, not even lateral control behaviors can be inferred as either the entire road surface is segmented instead of specific lanes or no semantic information about the identified lane markers is stored, respectively. To the best of our knowledge, there does not exist an automatic GT generation approach with which the specific kind of data that is required for the AI ego corridor model presented in \cite{Michalke2021} can be created. 
%
%In contrast to the previously detailed approaches, Rueden et al. \cite{Rueden2021} did not use an a-priori street map to generate a GT or estimate the position of lanes but rather validated a given segmentation mask by overlaying the segmentation with the course of the road extracted from the map. Applying this method to (automatically generated) GT data makes it possible to have an automatic and low-effort resource of annotation quality assurance. 

Summarizing the known literature in the area of automatic GT generation in the lane detection domain, \tref{Tab:StateofRes} shows that our work contributes the following new aspects compared to others:

\begin{itemize}
	\item Support of extended semantics beyond a drivable freespace: Labels for the ''AI ego corridor'' that enable lateral and longitudinal control,
	\item Explicit consideration of static and dynamic objects and occlusions that restrict the drivable corridor,
	\item Road type independent and compatible with marked and unmarked roads,
	\item Support of complex inner-city scenarios,
	\item Close coupling to a full-fledged L3 AD stack.
\end{itemize}

Major advantages are:
\begin{itemize}
        \item Holistic GT generation approach that supports all possible camera types given their intrinsics and extrinsics,
        \item Significant reduction of manual labeling effort by factor of \textgreater150.
\end{itemize}
% Downsides of current research:
% - GTs not as reliable and accurate in some cases (missing compensations/considerations, no HD maps used)
% - Often for highway scenarios (i.e. do not work on unmarked streets)
% - Different type of GT properties (e.g. only lane markers itself, approximated trajectory of street, cut-off at front of egolane (object detection), etc.)
% - Still require some amount of manual labeling (e.g. Time Slicing)

% distinction of different types of sensorsets?

% Potential Table Categories:
% X Oclusions considered (how?)
% - Exact position of lanes known or only course of road
% X Required type of GT (ego corridor instead of lane markig positions)
% X does it work on unmarked roads
% ? no human work required for GT *creation* (not including manual quality control)
% - constrained to specific area (due to HD map etc.)
% - purely vision-based
% - dynamic movement compensation (optimization with heuristic / based on sensor data)
% - range of approach

\begin{table*}[ht!]
  \caption{Selected state-of-the-art approaches for map-based automatic ground truth generation in the lane detection domain.}
  \begin{center}
    \begin{tabularx}{1.0\textwidth}{lllXXXXXX}%{@{}lllllllr@{}}
      \toprule[1.25pt]
      Method   & Year & Ref.               & Label Types & Ego Corridor GT Representation & Consideration of Objects and Occlusions & Compatible to Unmarked Roads + Inner-city scenes & Used Sensors Modalities \\ \midrule
      \'{A}lvarez et al.  & 2014 & \cite{Alvarez2014} & binary, pixel-wise road segmentation & - & implicitly & X & vision, GPS, GIS map
      \\
      Behrendt et Witt  & 2017 & \cite{Behrendt2017} & binary, pixel-wise segmentation of lane markings & - & -  & - & vision, GPS, HD map
      \\
      Behrendt et Soussan  & 2019 & \cite{Behrendt2019} & pixel-wise segmentation of individual lane markings and their semantic association & - & -  & - & vision, LiDAR, radar, odometry, GPS, HD map
      \\
      Kasmi et al.  & 2020 & \cite{Kasmi2020} & polynomial ego-lane representation & - & - & - & vision, LiDAR, odometry, GPS OpenStreetMap
      \\
      \midrule
      \textbf{Our approach}  & & & \textbf{binary pixel-wise segmentation of drivable ego-corridor} & \textbf{X} & \textbf{X} & \textbf{X} & \textbf{vision, LiDAR, odometry, GPS, HD map} \\
      \bottomrule[1.25pt]
    \end{tabularx}
  \end{center}
  \label{Tab:StateofRes}
\end{table*}

\section{Method}             \label{chap:method}

\begin{figure*}[htb]
	\centering
	\includegraphics[width=0.8\linewidth]{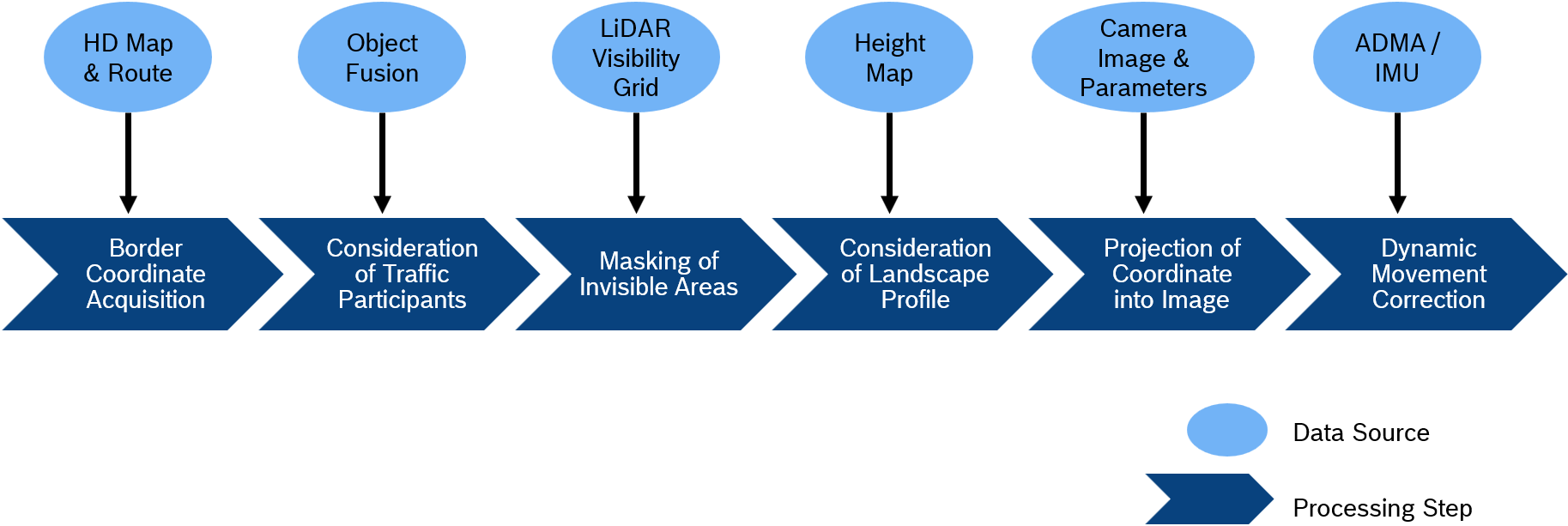}
	\caption{Data flow and processing sequence for automatic GT generation}
	\label{fig:pipeline}
\end{figure*}

The following section describes the proposed automatic GT generation method in detail, especially elaborating on which individual processing steps are performed to obtain an increasingly detailed and accurate GT representation. Each of those steps augments and enhances the desired GT instances with more information on the vehicle's current state and its surroundings to best capture the complexity and diversity of real-world traffic scenarios. These steps, together with the sequence in which they are executed and their respective information sources, are depicted in \fref{fig:pipeline}. Owing to its tight integration with the AD system as introduced in \fref{fig:ADpipeline}, this processing pipeline can be flexibly adjusted to generate alternative GT representations, to work with different sensor combinations, or to utilize additional information resources for specific scenarios. 
The following sub-sections discuss the individual processing steps in more detail.
\subsection{Border Coordinate Acquisition}
As the generated GT annotations are supposed to classify and indicate the areas in a camera image that belong to the current lane, the availability of accurate and suitable data on the positioning of the lane within the surrounding environment of the vehicle is the most essential part of this endeavor. Specifically, the course of the left and right lane boundaries, subsequently also referred to as \emph{lines}, are required to construct the corresponding corridor polygon. 

\begin{figure}[!htb]
	\centering
	\includegraphics[width=1.0\columnwidth, trim={3.0cm 0.0cm 4.0cm 2.25cm},clip]{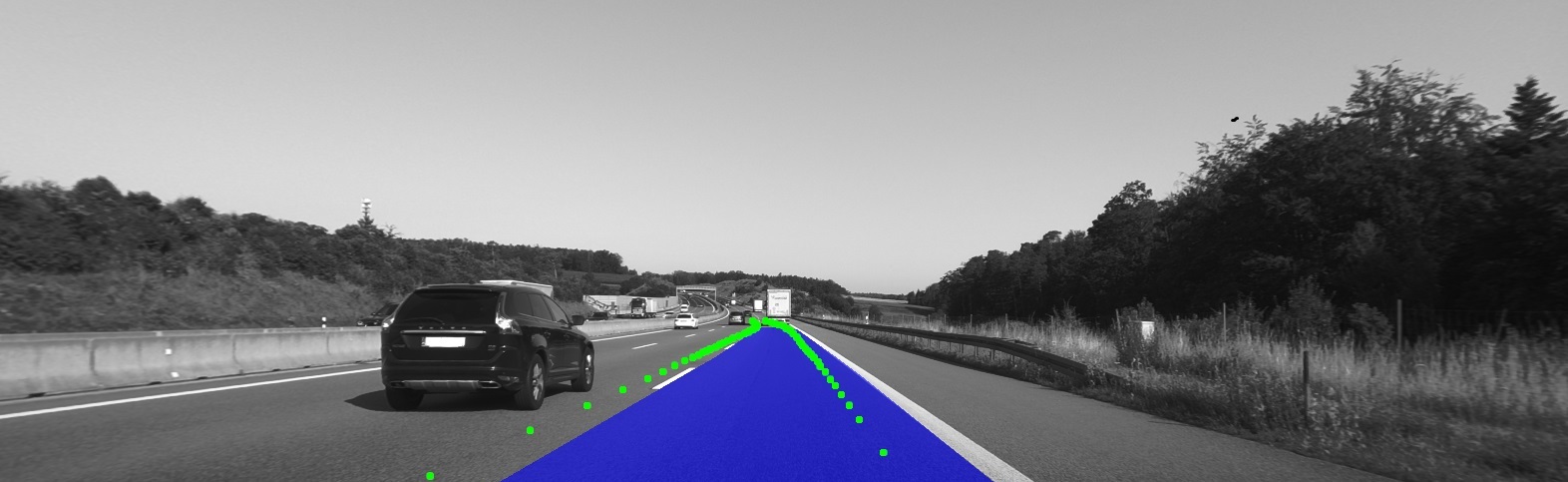}
	\caption{Effects of shifting original map-based corridor coordinates (green) according to online lane detection system to obtain better fit (blue). The error was induced by a local ego-localization inaccuracy.}
	\label{fig:lines-dynamicShift}
\end{figure}

Utilizing HD maps from which the lines can be extracted has several advantages. As the course and position of the road can be inferred directly from the map, the range in which this approach can operate is practically unlimited. Additionally, even in challenging environmental situations or with objects obstructing parts of the lane, highly accurate and detailed line information are still available. This HD map-based approach is only limited by the precision of the ego-localization module that determines the vehicle's current position within the environment and map. Through the alignment of map-based and online lane detections, for example, localization inaccuracies and errors can be significantly mitigated as illustrated in \fref{fig:lines-dynamicShift}.
\subsection{Consideration of Traffic Participants}
In order to accordingly adjust the corridor such that it does not interfere with and overlap any traffic participant occupying parts of the own lane, information on traffic participants determined by the object fusion module of the vehicle system is used. Here, data and predictions from video, radar and LiDAR object detection algorithms are combined and tracked, leading to highly accurate and dependable object classification and property assessment. Besides the position of an individual object with respect to the ego-vehicle, the relative velocity and the direction of travel are also ascertained. The classes of objects and traffic participants that are distinguished and detected by the module include motorized vehicles like cars, trucks and motorcycles, non-motorized road users in the form of pedestrians and cyclists, as well as stationary objects like cones and bollards.

\begin{figure}[!htb]
	\centering
	\resizebox{\columnwidth}{!}{
		\begin{tikzpicture}
		\clip (-3.5,5.5) rectangle + (19+3.5,-5.5-6.5); 
		
		\draw[double=white,double distance=2.5cm,ultra thick] (0,-5) to[in=-180, out=90, looseness=1] (18,4) -- (-2,4); 
		\draw[thick,dashed,dash pattern=on 20pt off 20pt] (0,-5) to[in=-180, out=90, looseness=1] (18,4); 
		\draw[double=blue!20,double distance=1.15cm] (0.64,-5) to[in=-180, out=90, looseness=0.975] (18,3.35); 
		\draw[thick,dashed,dash pattern=on 20pt off 20pt] (-2,4) -- (10,4); 
		\node[inner sep=0pt,rotate=-90,scale=0.55, label={[label distance=0.5cm]\Huge Ego-vehicle}] (whitehead) at (0.7,-5) {\includegraphics[width=.25\textwidth]{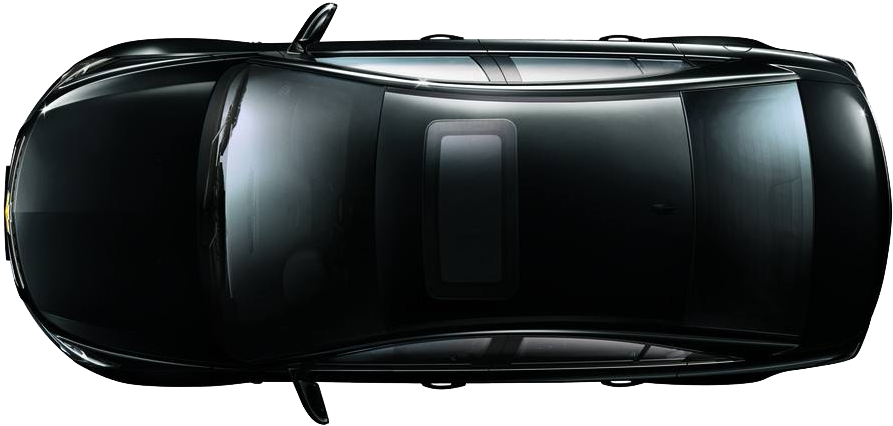}}; 
		\node[inner sep=0pt,rotate=-180,scale=0.55, label={[label distance=0.3cm, xshift=0.6cm]-90:\Huge Object A}] (whitehead) at (15,3.3) {\includegraphics[width=.25\textwidth]{pictures/car_topview.png}}; 
		\node[inner sep=0pt,rotate=-200,scale=0.55, label={[label distance=0.5cm]-70:\Huge Object B}] (whitehead) at (7.8,3.0) {\includegraphics[width=.25\textwidth]{pictures/car_topview.png}}; 
		
		\draw[thick,red] (13.95,1.5) -- (13.8,5); 		
		\node[label={[label distance=0.5cm, xshift=0.5cm, red]\huge Cutoff line A}] (a) at (14.5,0) {}; 
		\node[label={[label distance=0.5cm, xshift=-0.5cm, red]\huge Cutoff line B}] (a) at (11,0) {}; 
		\draw[thick,red] (10.5,1.5) -- (5,3.5); 
		\draw[-stealth, black, line width= 1.4mm] (7.5,3.1) -- (9.5,2.4); 
		\draw[-stealth, black, line width= 1.4mm] (15,3.3) -- (17,3.3); 

		\end{tikzpicture}
	}
	\caption{Corridor cut-off scenarios for objects and traffic participants}
	\label{fig:objcts-overview}
\end{figure}

While there might be many traffic participants and objects in the surrounding environment of the ego-vehicle, most of them are irrelevant and do not need to be considered for the GT generation process. Only objects adjoining and intersecting with the corridor are of importance. For traffic participants that meet these requirements, two different scenarios are distinguished (see \fref{fig:objcts-overview}): traffic that uses the same stretch of road and follows the same general route as the ego-vehicle (object A) and cross-traffic that intersects or crosses with the ego-lane (object B). The distinction between these scenarios is necessary as the corridor has to be cut off differently depending on how the objects intersect with the corridor such that the side closest to the corridor is taken as the cut-off line.

\begin{figure}[!ht]
	\centering
	\subfloat[Traffic within own lane \label{fig:objcts-scenario1}]{\includegraphics[width=0.49\columnwidth]{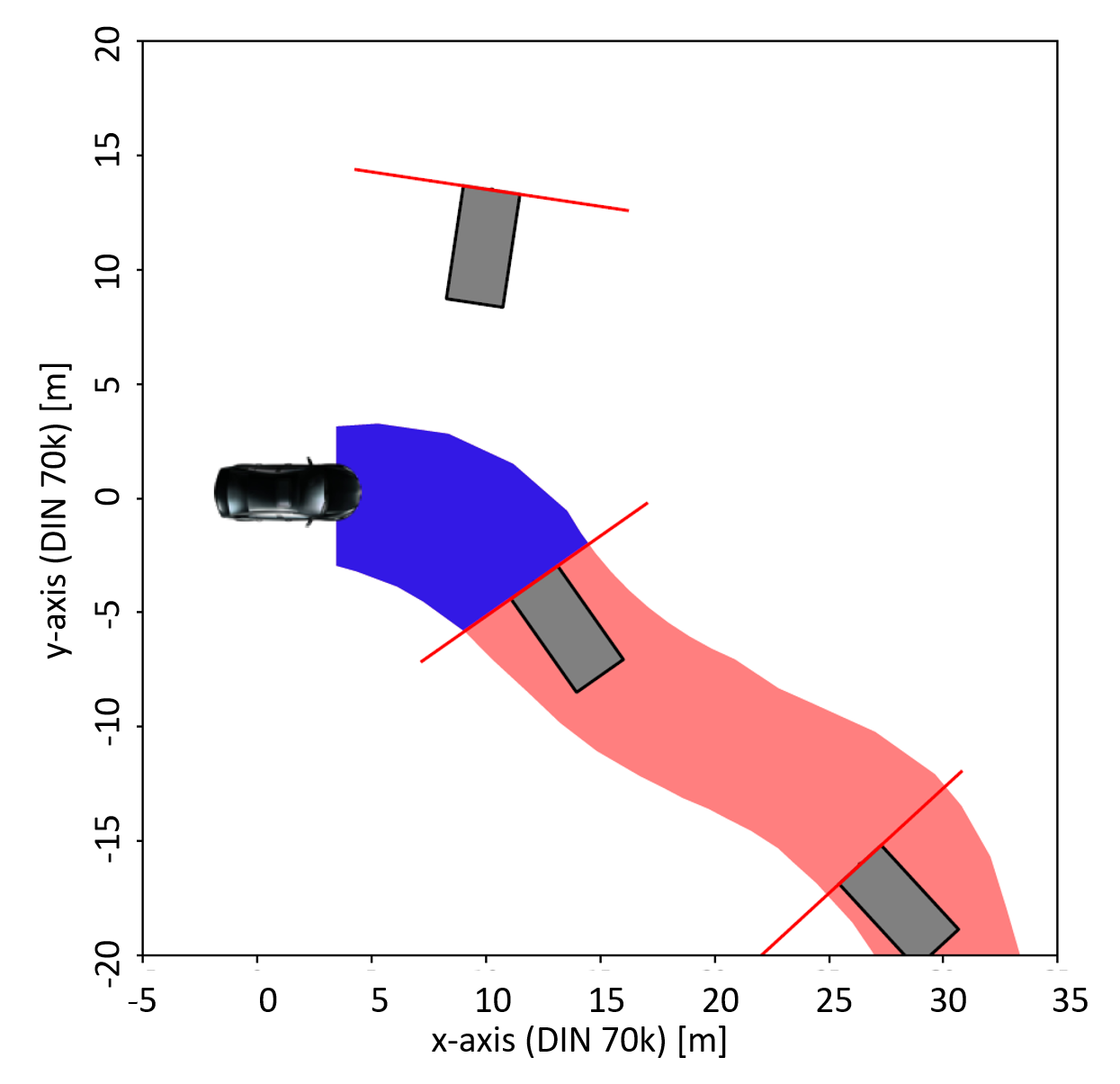}}\hfill
	\subfloat[Traffic crossing own lane \label{fig:objcts-scenario2}]{\includegraphics[width=0.49\columnwidth]{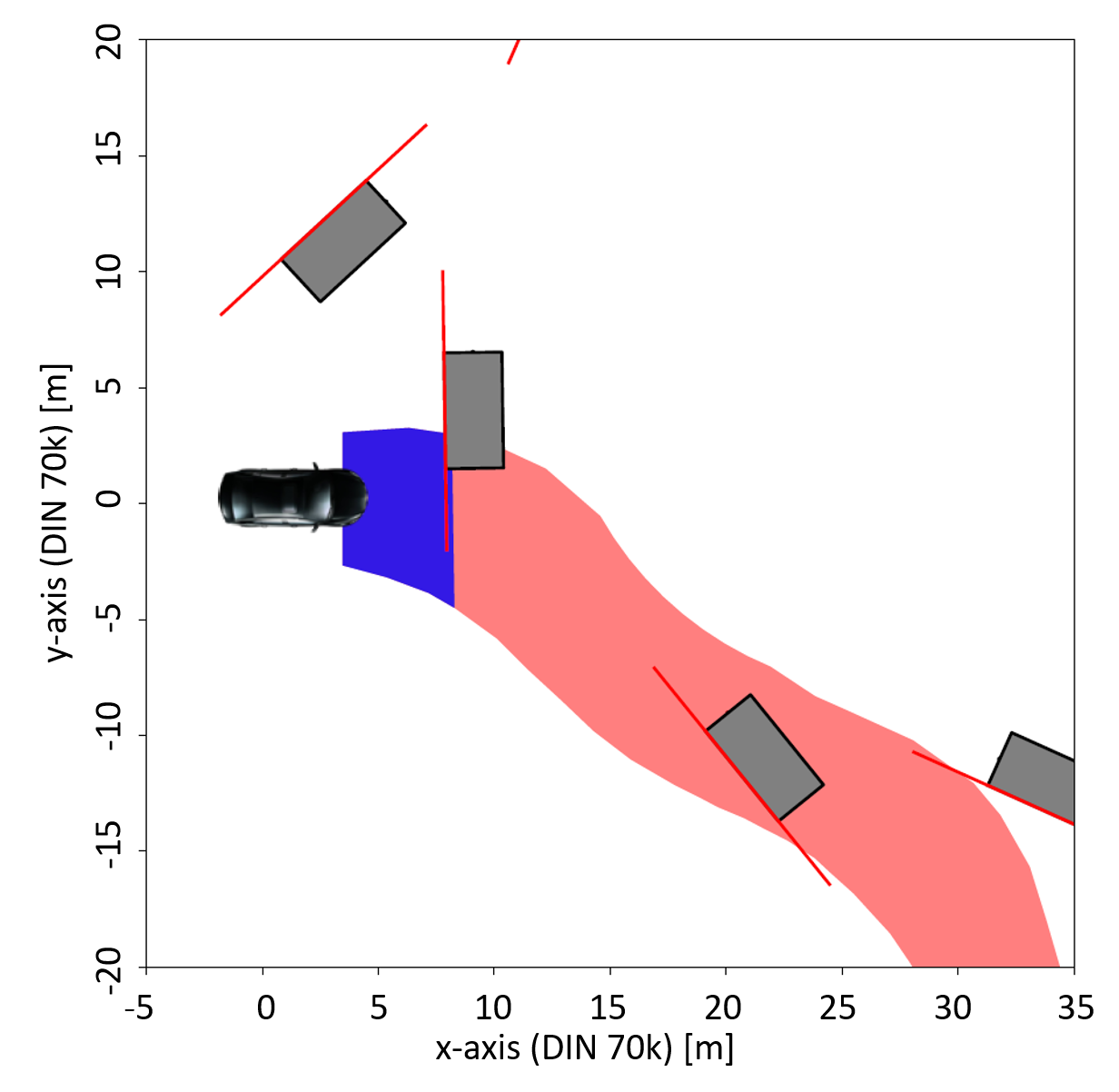}}
	\caption{Modifications to the original corridor in case of traffic participants}
	\label{fig:objcts-scenarios}
\end{figure}

For traffic driving ahead, the own corridor needs to be cut off accordingly such that it ends right before the preceding traffic participant. For this purpose, the corridor is split along a line angled perpendicular to the object's direction of travel and originating from the rearmost point of the object. By taking the resulting part of the split operation closest to the ego-vehicle, the new corridor instance can be obtained. In case of the cross-traffic scenario, the corridor is modified similarly, but the cut-off line originates not from the rearmost point but from a point of the bounding box's side facing the corridor instead and shares the same orientation as the object. Figures \ref{fig:objcts-scenario1} and \ref{fig:objcts-scenario2} each illustrate the original corridor (red), the position of objects relative to the ego-vehicle and the accordingly cut-off new corridor instance (blue) for both of the introduced scenarios respectively.
\subsection{Masking of Occluded Areas}
While the corridor is already cut off in case of any present traffic participants within the own lane at this stage, other obstacles that possibly obstruct the camera's view are not considered yet. Especially in urban driving scenarios, however, obstacles like buildings, parked cars or vegetation can cause the direct line of sight from the camera to the corridor to be blocked. Therefore, it is especially important for these areas to be identified and removed from the corridor representation accordingly. 

\begin{figure}[!ht]
	\centering
	\subfloat[Visibility grid not considered \label{fig:visGrid-before}]{\includegraphics[width=1.0\columnwidth]{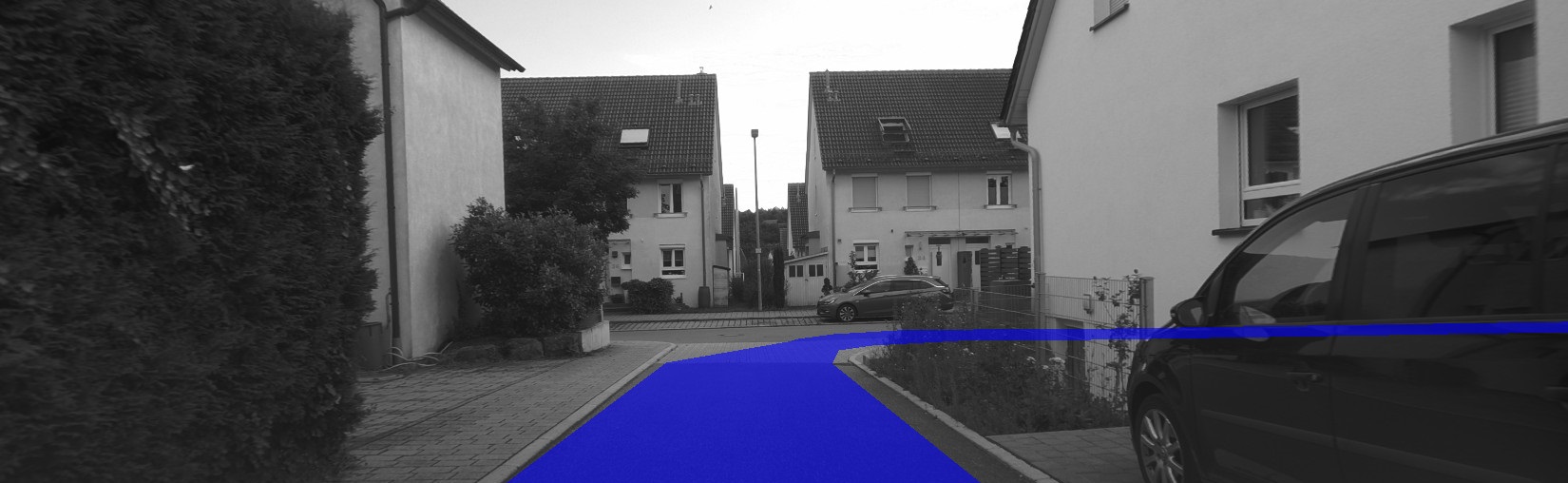}}
	
	\subfloat[Visibility grid considered \label{fig:visGrid-after}]{\includegraphics[width=1.0\columnwidth]{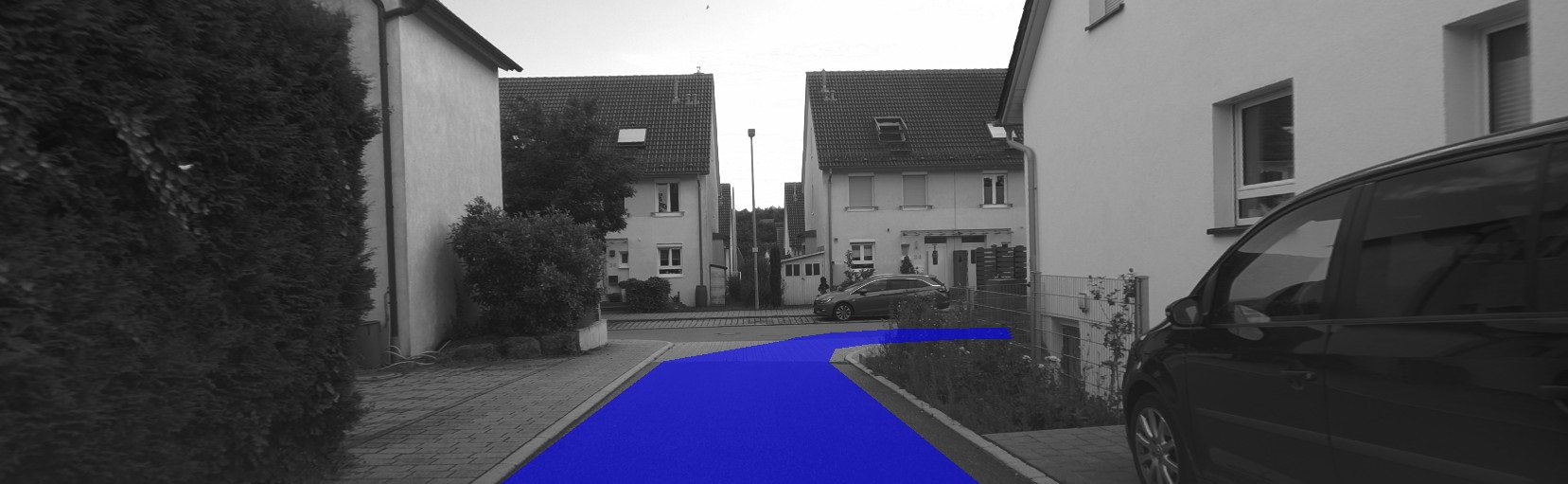}}\hfill
	\caption{Effects of modifying the corridor according to the LiDAR-based visibility grid}
	\label{fig:visGrid-comparison}
\end{figure}

Contrary to the dynamic object's removal workflow, here, areas not visible to the camera are only stamped out instead of cutting off the corridor completely. In order to obtain information about these areas, a visibility grid constructed on the basis of LiDAR-sensor data is employed. This grid is constructed by subdividing the world around the vehicle into distinct cells and computing the cost of traversing the specific cells to obtain if an obstacle is present \cite{ROSWC2018}. This three-dimensional representation, the so-called \emph{costmap}, of the surrounding environment is then projected into a two-dimensional binary map that either denotes the existence of an obstruction or the possibility of unobstructed view and movement. By assuming the areas behind obstacles to be occluded and not visible in camera images, the intended visibility grid can be obtained. It is now possible to determine which parts of the drivable corridor are obstructed and which ones are not. When applying this processing procedure to an urban scenario, parts of the initially projected drivable corridor corresponding to the obstructed areas are removed as depicted in \fref{fig:visGrid-comparison}.
\subsection{Consideration of Landscape Profile} \label{sec:elevation_profile}
Contrary to the flat world assumption in which the environment is regarded as an even two-dimensional surface with no differences in height throughout the plane shared among the previous processing steps, the landscape, in reality, is shaped by — sometimes significant — variations in altitude and slope. Therefore, it is essential to compensate and consider the effects of changes in the landscape profile, if an accurate projection is desired.

\begin{figure}[!ht]
	\centering
	\subfloat[Height map not utilized \label{fig:height-before}]{\includegraphics[width=1.0\columnwidth]{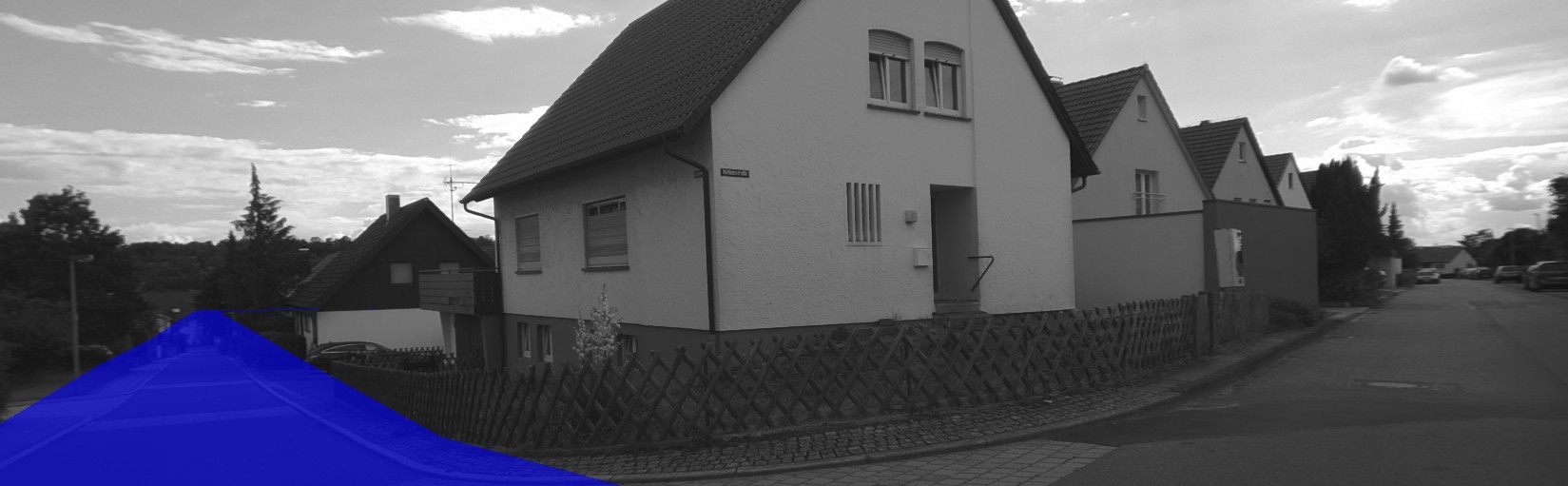}} \hfill
	\subfloat[Height map utilized \label{fig:height-after}]{\includegraphics[width=1.0\columnwidth]{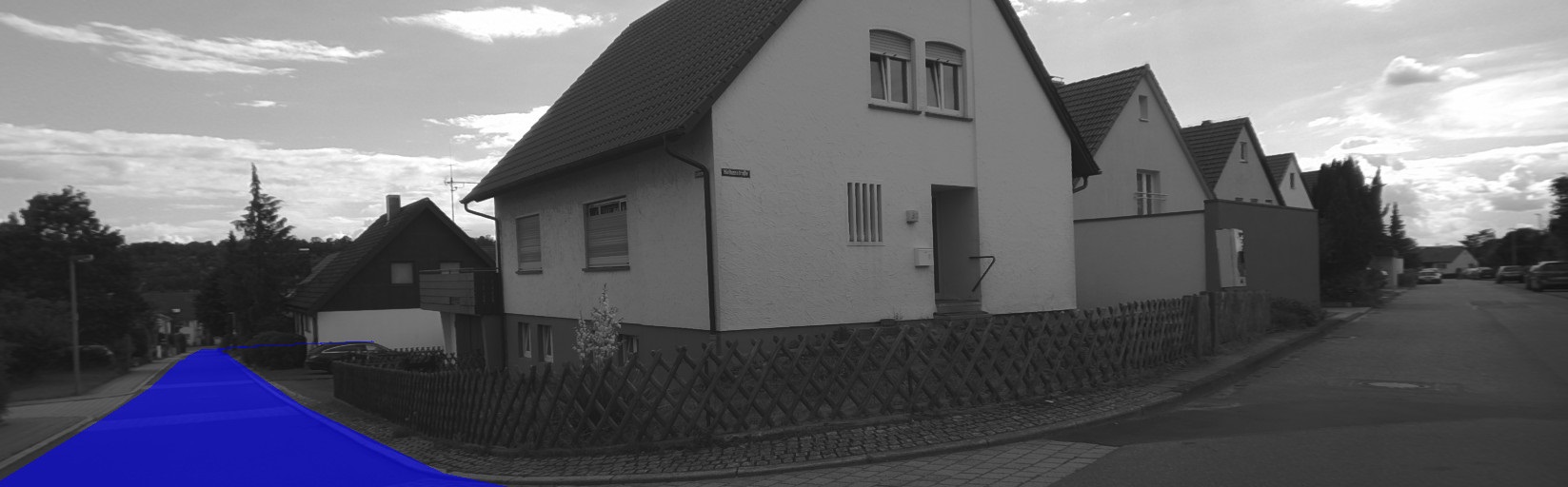}}
	\caption{Effects of height consideration on corridor projection in urban scenario with notable changes in elevation.jpg}
	\label{fig:height-comparison}
\end{figure}

As the utilized HD map does not incorporate precise altitude data, it is necessary to construct a height map along the driven route specifically. By recording the altitude measurements of the localization system and combining them with the associated measured position of the vehicle, such a height reading can be mapped to a unique and unambiguous location. Different from online application of the AD stack, the GT generation can realize an ex-post analysis, running the AD systema recorded datasets and gather the altitude and corresponding positional data. Subsequently, these measurements can then be stitched together into a map of the desired format which can be acted upon in future GT generation runs to approximate the altitude of arbitrary points along the route. The originally two-dimensional corridor representation can thus be iteratively transformed to also incorporate elevation data and constitute a three-dimensional representation of the drivable area. Thereby, it is eventually possible to substitute a projection like in \fref{fig:height-before} with a significantly more accurate annotation as \fref{fig:height-after} demonstrates.
\subsection{Corridor Projection}
While the corridor has been available and modified in its 3D representation relative to a vehicle reference point so far, it is ultimately necessary to project the relevant areas onto the actual image. For this translation, a pinhole camera model is employed. Given calibrated intrinsic and extrinsic camera parameters, this model allows a world position to be translated into the corresponding pixel coordinates of any camera system mounted on the vehicle. Doing so for all positions spanning the drivable corridor eventually yields corresponding drivable areas within the image plane. After these steps, projections like already depicted in Figures \ref{fig:visGrid-comparison} or \ref{fig:height-comparison} with the drivable areas clearly and accurately designated can be obtained. 
\subsection{Dynamic Movement Correction}
\begin{figure}[ht]
	\centering
	\subfloat[No tilt compensation \label{fig:tilt-before}]{\includegraphics[width=1.0\columnwidth]{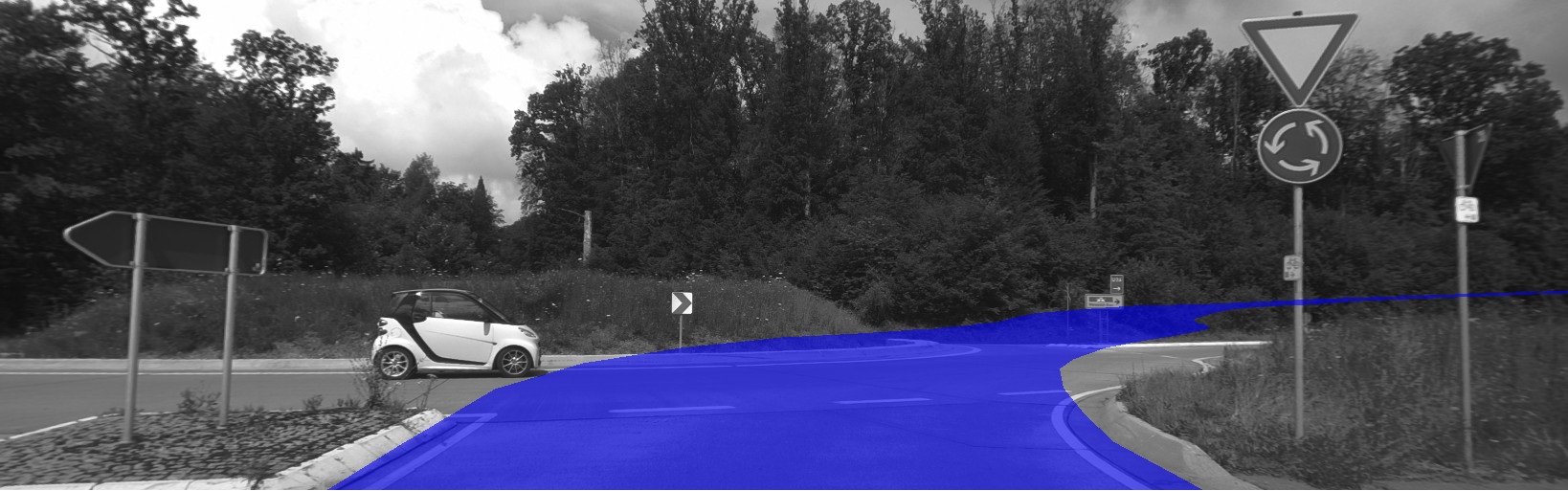}}
	
	\subfloat[Active tilt compensation \label{fig:tilt-after}]{\includegraphics[width=1.0\columnwidth]{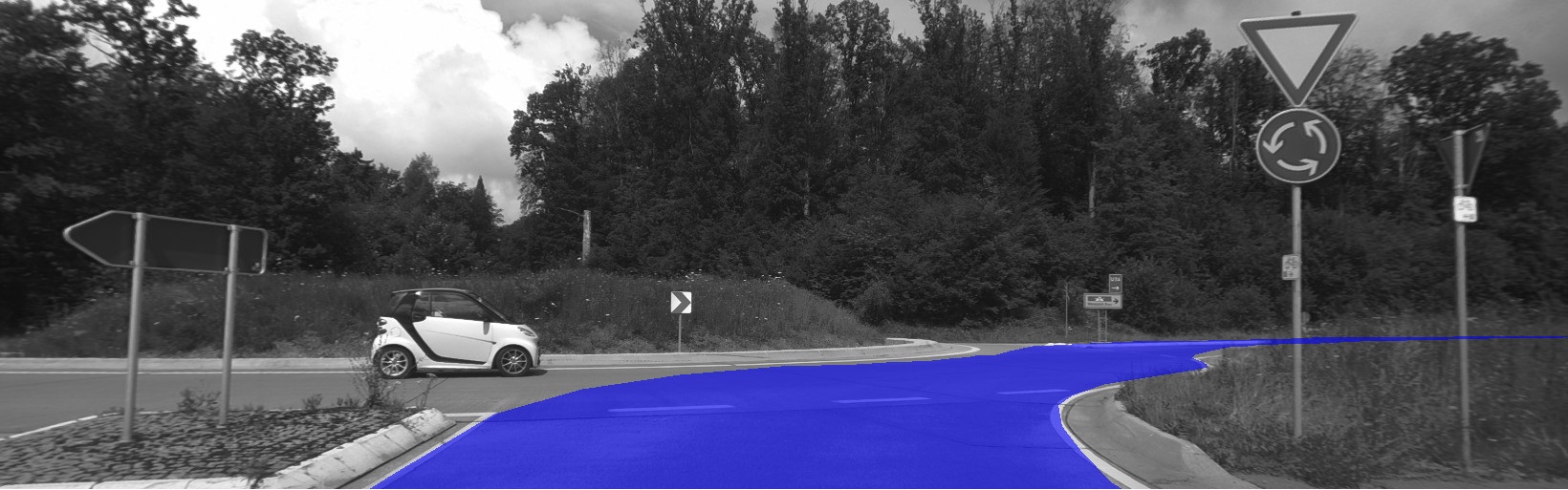}}
	\caption{Impact of compensating dynamic vehicle movements and road gradients}
	\label{fig:tilt-comparison}
\end{figure}

As a dynamic system that is affected by various external forces and influences, the ego-vehicle's positional state and, as a result, the camera's position and orientation are constantly changing. This might be due to inclines and slopes of the road, potholes and other surface irregularities, or varying pitch and roll angles through changes in inertia and centrifugal forces. Such variations need to be considered and compensated for the projected corridor to match the corresponding and appropriate image positions. Besides compensating movement brought about by these effects, it is also necessary to consider the current vehicle tilt when utilizing the height map approach from \sref{sec:elevation_profile}. Both of these effects have an impact on the projection in \fref{fig:tilt-before} but can be compensated through according measures to obtain the accurate projection in \fref{fig:tilt-after}.

In particular, changes in the roll and pitch angle are relevant for this process as even minor deviations in the orientation of the camera can result in substantial displacements for faraway objects and landmarks \cite{Behrendt2019}. In contrast, slight changes in the effective camera height above the ground caused by, for example, the compression of the suspension only have a minor impact on the projection and occur to a considerably smaller extent in typical driving scenarios. The roll and pitch angles of the vehicle are measured by the DGPS-supported innerial measurement unit of the AD system and are compensated by altering the projection process through the addition of an upstream modification of the world coordinates before being plugged into the camera matrix model. Specifically, this is done to rotate the employed coordinate system such that it still matches the camera's orientation in the real world. Thus, a separate rotation matrix is constructed that rotates the input positions around the roll-axis followed by the pitch-axis, compensating dynamic variations.

\section{Experiments}      \label{chap:Experiments}

In the following section, qualitative and quantitative measures are employed to assess both the automatically generated GT annotations as well as a ML model trained therewith.

For the quality of a generated GT annotation to be conclusively assessed, it was necessary to compare it against a manually created annotation representing the segmentation mask's desired target stage. Here, highway and more challenging urban driving scenes are separately analyzed. Owing to the laborious nature of the manual labeling process, only a total of 20 batches, each containing 25 images, were randomly selected out of all available image-mask pairs amounting to a total of 500 images for each scene type to be manually annotated. With a corpus of approximately $12,000$ and $11,000$ GT instances that passed basic quality assurance measures for highway and urban scenarios, respectively, this sampling strategy still retains a representative, dependable and unbiased assessment of the approach's performance.

Using metrics that measure the misalignment between automatically ($aGT$) and manually created ($mGT$) segmentation masks, it is possible to quantitatively judge and assess the quality of the generated GT labels. As common practice in the semantic segmentation space \cite{Taha2015}, it was opted to utilize the \emph{Dice coefficient} (\eref{formula:Dice}) \cite{Dice1945} and the Jaccards index (\eref{formula:Jac}) \cite{Jaccard1912} to ascertain the differences. 
\begin{equation}
\mathit{DICE} = \frac{2 |aGT \cap mGT|}{|aGT| + |mGT|}
\label{formula:Dice}
\end{equation}
\begin{equation}
\mathit{JAC} = \frac{|aGT \cap mGT|}{|aGT \cup mGT|}
\label{formula:Jac}
\end{equation}
With a mean of both metrics of $98.1\%$ in the case of the highway scenarios, the general quality of the generated GT annotations accurately corresponds to the manually created masks. Only slight variations in the score across the individual sample batches suggest consistent qualitative characteristics in different driving and environmental scenarios. 

This provides preliminary evidence of the approach's suitability for a successful deployment which is further reinforced by the encouraging albeit somewhat lower scores in more diverse and complex urban scenarios. Here, an overall score of $94.0\%$ is reached. Dissecting the urban sample batches into further types of scenarios for a more expressive explanation of the underlying effects, as seen in \tref{tab:resultsUrban}, shows that especially parking cars negatively impact the quantitative judgment. 

\begin{figure}[!ht]
	\centering
	\subfloat[Cut-off corridor due to overly extended object bounding box \label{fig:parkingIssue1}]{\includegraphics[width=0.49\textwidth, clip]{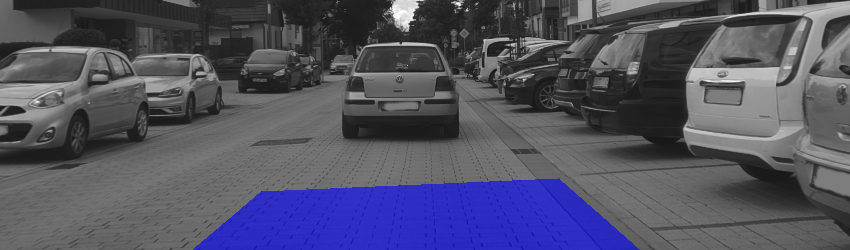}}\hfill
	\subfloat[Cut-off corridor due to parking car within ego-lane \label{fig:parkingIssue2}]{\includegraphics[width=0.49\textwidth, clip]{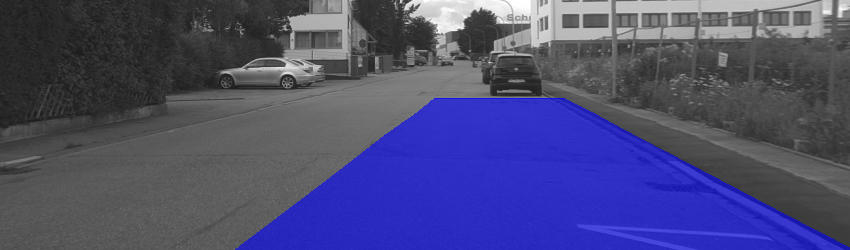}}
	\caption{Issues evoked by insufficient consideration of parked vehicles}
	\label{fig:parkingIssue}
\end{figure}

\begin{figure}[!ht]
	\centering
	\subfloat[Generated GT image \label{fig:curveIssue1}]{\includegraphics[width=0.49\textwidth,clip]{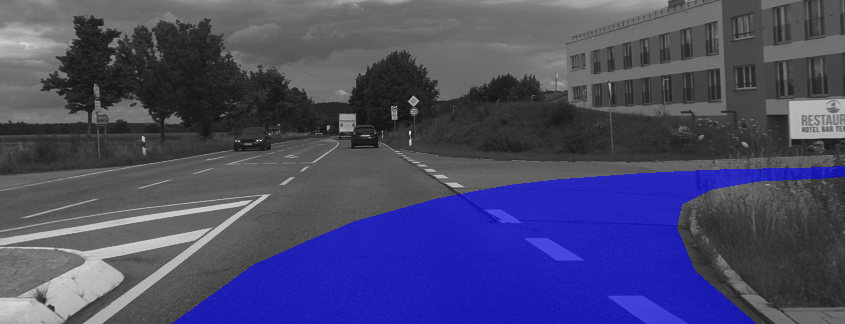}}\hfill
	\subfloat[Human annotation \label{fig:curveIssue2}]{\includegraphics[width=0.49\textwidth,clip]{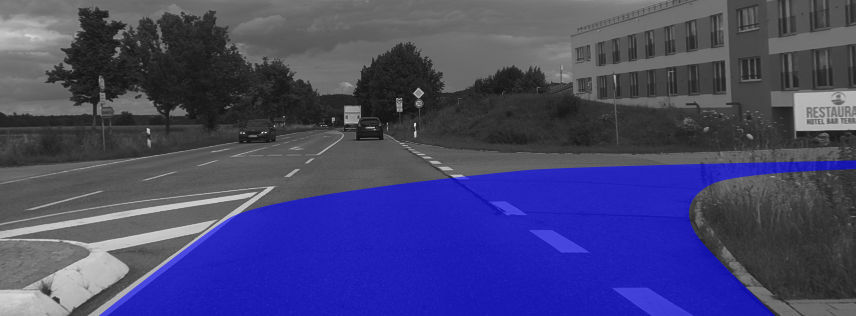}}
	\caption{Difference in the proposed route during sharp turn on unmarked road}
	\label{fig:curveIssue}
\end{figure}

In some cases, this effect is due to inaccurate placements of detected objects resulting in them being mistakenly cut off in case vehicles are parked in proximity to the current lane (see Figure \ref{fig:parkingIssue1}). Another related problem is that parked, and moving cars are not robustly distinguished. Therefore, the corridor is cut off right in front of them instead of detouring around them as depicted in Figure \ref{fig:parkingIssue2}. These problems could be solved by classifying and discerning mobile and immobile objects and cutting off the corridor or removing the intersection with the corridor, respectively. 
\begin{table}[!htb]
	\centering
	\small
        \vspace{0.5cm}
	\begin{tabular}{lcccc} \toprule
		\textbf{Scenario} & \textbf{\#} & \textbf{$\mathit{DICE}$} & \textbf{$\mathit{JAC}$} & \textbf{$\mathit{AVG}$} \\ \midrule
		\textbf{Sharp Curve} & 6 & 0.934 & 0.891 & 0.913 \\ \hdashline
		\textbf{No Markings} & 7 & 0.974 & 0.960 & 0.967 \\ \hdashline
		\textbf{Parking Cars} & 3 & 0.880 & 0.838 & 0.859 \\ \hdashline
		\textbf{Others} & 4 & \textbf{0.997} & \textbf{0.994} & \textbf{0.995} \\ \midrule
		\textbf{All (weighted)} &  & 0.953 & 0.928 & 0.940 \\ \bottomrule
	\end{tabular}
	\caption{Performance of the GT generation approach on urban sequences}
	\label{tab:resultsUrban}
\end{table}
Aside from these effects, the quality of the other scenarios was reasonably good, however. Namely on road segments without lane markings and in the ''Others'' category (includes mainly inter-urban passages with clear lane markings and no significant curvature), the attained scores demonstrated a solid segmentation accuracy. The reduced quality during sharp curves and junctions is, to a large extent, caused by sequences during which no clear lane markings or directional references are available. One example of such an instance that was part of the evaluation dataset is illustrated in Figure \ref{fig:curveIssue}. Whereas the generated mask directs the route more straightforwardly towards the diverging street, the manual annotation intends for the turning maneuver to be later and therefore covers a considerably greater ego-lane area at the given moment, leading to a penalization that is reflected in the scores. Due to the lack of conclusive markings, it is hard to judge, which alternative is more accurate, reducing the expressiveness of the utilized metrics in such scenarios.

Summarizing these results, the obtained scores suggest that the automatically created GT annotations - despite not quite matching their human-created counterparts across all situations - are of sufficient accuracy to be used to train ML models. In addition, higher degrees of labeling noise can be compensated by the sheer volume of generatable GT data. This can be seen in \ref{tab:resultshighway}, where a network trained on manual labelled data is compared to a network trained on automatically generated data. We used the highway network architecture proposed in \cite{Michalke2021} and unseen hand-labeled frames for the KPI generation.
\begin{table}[!htb]
	\centering
	\small
	\begin{tabular}{lcccc} \toprule
		\textbf{GT type} & \textbf{\# of labels} & \textbf{manual effort} & \textbf{$\mathit{DICE}$} & \textbf{$\mathit{JAC}$} \\ \midrule
		\textbf{\textbf{Auto-generated}} & \textbf{20,000} & \textbf{15min} & \textbf{0.884} & \textbf{0.936} \\  \midrule
		\textbf{Manual} & 5,000 & 11h & 0.869 & 0.923  \\ \bottomrule
	\end{tabular}
	\caption{Comparison of model performance trained with manual and auto-generated GT on highway sequences}
	\label{tab:resultshighway}
        %\vspace{-1cm}
\end{table}

\section{Conclusion and Outlook}
The results presented in the previous section provide convincing evidence that the proposed GT generation approach does yield GT annotations that nearly match the quality of their manually created counterparts. Only requiring basic quality assurance of 15 minutes or less to remove obviously flawed instances for the generation of 20,000 GT annotations, our developed approach vastly outperforms the human labeling process requiring approximately 40 hours of work for a dataset of the same magnitude. In our experience, deploying the automatic GT can result in time savings by a factor of over 150. 

As a result, significantly larger volumes of GT data are generatable with only minimal human intervention. Subsequently, ML models with more complex architectures and better as well as more robust performance should be trainable. The availability of training and evaluation data is, therefore, no longer the bottleneck of the AI corridor approach, making it much more practicable and supporting its deployment in the real world. 

In the future, we plan to realize an automated data-loop and want to research its impact on the network performance.

\addtolength{\textheight}{-12cm}   % This command serves to balance the column lengths

%%%%%%%%%%%%%%%%%%%%%%%%%%%%%%%%%%%%%%%%%%%%%%%%%%%%%%%%%%%%%%%%%%%%%%%%%%%%%%%%
%References are important to the reader; therefore, each citation must be complete and correct. If at all possible, references should be commonly available publications.
%working for git version
\bibliographystyle{./bibtex/IEEEtran} % use IEEEtran.bst style
\bibliography{./bibtex/IEEEabrv,./bibtex/paper}

\end{document}